\documentclass[11pt]{article}
\usepackage[hyperref]{eacl2021}
\usepackage{times}
\usepackage{latexsym}

\usepackage{microtype}

\usepackage{times}
\usepackage{amsmath}
\usepackage{calc}
\usepackage{latexsym}
\usepackage{booktabs}
\usepackage{tikz}
\usepackage{tikz-qtree}
\usetikzlibrary{decorations.text, decorations.pathreplacing}
\usepackage{url}
\usepackage{multirow,rotating}
\usepackage{booktabs,array,xcolor,makecell}
\usepackage{microtype}
\usepackage{hyphenat}
\usepackage{enumitem}
\usepackage[scaled=0.89]{helvet} 

\usepackage{float}
\restylefloat{table}

\aclfinalcopy 

\newcommand{\ecframe}[1]{\textbf{#1}}

\newcommand{\ocrcomment}[1]{}

\newcommand{\sesame}{Open Sesame}
\newcommand{\fullgen}{Full-Gen}
\newcommand{\multitask}{Multi-Task}

\newcommand{\exactmatch}{Exact Match}
\newcommand{\softmatch}{Soft Match}
\newcommand{\microsoftmatch}{Global Match}

\title{Open-Domain Frame Semantic Parsing Using Transformers}

\author{Aditya Kalyanpur, Or Biran, Tom Breloff, Jennifer Chu-Carroll, \\
\textbf{Ariel Diertani},
\textbf{Owen Rambow}\Thanks{Current affiliation: Stony Brook University}, \textbf{Mark Sammons} \\
  Elemental Cognition\\
  New York, NY, USA\\
  \texttt{\{adityak, orb, tomb, jenniferc, arield,} \\ \texttt{owenr, marks\} @elementalcognition.com}\\\vspace{-1cm}
  }

\date{}

\begin{document}
\maketitle
\begin{abstract}
Frame semantic parsing is a complex problem which includes multiple underlying subtasks. Recent approaches have employed joint learning of subtasks (such as predicate and argument detection), and multi-task learning of related tasks (such as syntactic and semantic parsing). In this paper, we explore multi-task learning of all subtasks with transformer-based models. We show that a purely generative encoder-decoder architecture handily beats the previous state of the art in FrameNet 1.7 parsing, and that a mixed decoding multi-task approach achieves even better performance. Finally, we show that the multi-task model also outperforms recent state of the art systems for PropBank SRL parsing on the CoNLL 2012 benchmark.
\end{abstract}

\section{Introduction}
The problem of frame semantic parsing \cite{gildea-jurafsky-2000-automatic} has shown value in numerous NLP applications from dialog systems \cite{chen2013unsupervised} to question answering \cite{shen2007using}. More recently, semantic parsing has been used in interactive chat-bots such as Google's Dialogflow or Amazon's Alexa skills to map the user's language (command) to a semi-formal structure that captures the intended task, often represented as frames with fillable slots \cite{chen2019bert}

Frame semantic parsing is often formulated as a set of underlying tasks which are highly dependent on one another. For example, FrameNet \cite{framenet:book} parsing is modeled as a sequence of
trigger selection, frame labeling, argument span selection and argument labeling. In previous work, the tasks have been tackled in a pipeline fashion, with the (true or predicted) value of one limiting the possibilities of the next \cite{das_etal_2014,swayamdipta_etal_2017}. 
Recently, pre-trained transformer-based neural models \cite{attention2017} have achieved state-of-the-art performance on various NLP problems \cite{devlin2018bert,gpt2,2019t5}. In this paper, we explore the application of transformer-based architectures to frame semantic parsing. Additionally, we take advantage of the relatedness of the underlying tasks by employing multi-task learning, and show significant improvement over previous state-of-the-art results. 

In our most straightforward approach, we treat semantic parsing as a sequence-to-sequence generation problem, 
where the input is natural language text and the output is a textual encoding of the corresponding frame representation. We show that this approach achieves higher performance when using an encoder-decoder architecture such as T5 \cite{2019t5} compared to a monolithic language model such as GPT2 \cite{gpt2}. Interestingly, and somewhat surprisingly, we find that this straightforward seq2seq formulation using the T5-base model shows a massive improvement (12-17\% on argument detection F1) on the FrameNet 1.7 evaluation benchmark, compared to a recent state-of-the-art semantic parser, Open-Sesame \cite{swayamdipta_etal_2017}, which highlights the strength of using pre-trained transformer models for this task. 

\begin{table}[ht]
    \centering
    \small
    \begin{tabular}{|p{2cm}|p{4.9cm}|}
        \toprule
        \textbf{Sentence with frame trigger underlined} & \textbf{Semantic Parse: Frame with \{arg : value ..\} assignments} \\\midrule
        Mary \underline{got} to the train station at 2pm. & \emph{Arriving} \{\textbf{theme}: Mary, \textbf{goal}: train station, \textbf{time}: 2pm\} \\\hline
        John \underline{got} an apple from his backpack. & \emph{Getting} \{\textbf{recipient}: John, \textbf{theme}: apple, \textbf{source}: backpack\} \\\hline
        Tom \underline{got} into trouble for that. & \emph{TransitioningToState} \{\textbf{entity}: Tom, \textbf{finalSituation}: trouble\} \\ \hline
        He \underline{ate} a sandwich since he was hungry & \emph{Ingestion} \{\textbf{ingestor}: He, \textbf{ingestibles}: sandwich, \textbf{explanation}: he was hungry\} \\ \hline
        He \underline{picked up} Math quite easily in school & \emph{Grasp} \{\textbf{cognizer}: He, \textbf{phenomenon}: Math, \textbf{manner}: quite easily\} \\ \hline
        She \underline{sneezed} the napkin off the table. & \emph{Removing} \{\textbf{agent}: She, \textbf{theme}: napkin, \textbf{source}: table\} \\
        \bottomrule
    \end{tabular}
    \caption{FrameNet Semantic Parsing Examples using our multi-task parser}
    \label{table:spindleExamples}
\end{table}

We also experiment with a traditional formulation of the problem which decomposes semantic parsing into frame classification (using a multi-class classifier) followed by argument detection (using a generative decoder) conditioned on the frame. For this purpose, we build a general multi-task learning architecture, where a common encoder is shared among the related tasks, and its output is fed to task-specific classifiers and generative decoders.
At prediction time, we run the model for each of the sub-tasks, gather the results, and compose the final frame interpretation. See Table \ref{table:spindleExamples} for examples of sentences parsed with our multi-task parser.
We show that a 2-task model that uses explicit token positional indices in its encoding achieves a further improvement of 1-5\% on role/argument F1 across various settings (of dev/test and predicted/gold predicates), over the pure generative seq2seq approach. 



This paper is structured as follows: Section~\ref{sec:related_work} presents an overview of related work. In Section~\ref{sec:hector}, we discuss the three frame ontologies used in this paper: the PropBank sense hierarchy and its semantic rolesets \cite{palmer-etal-2005-proposition}, the FrameNet ontology, and our broad-domain frame ontology called \emph{Hector}. Section~\ref{sec:models} describes our various model architectures and show experiments comparing the performance of different models on the Hector dataset (the larger and more expressive of the ontologies). We then show, in Section~\ref{sec:benchmark_experiments}, the performance of our best model on standard semantic parsing data sets, namely FrameNet and PropBank, and conclude.



\section{Related Work}
\label{sec:related_work}

Most recent work on frame semantic parsing has focused on joint or multi-task learning in one form or another. 
In joint learning, a single model is learned using a single loss function across multiple related tasks, so that results on any of the tasks are improved and unlikely or impossible combinations are avoided.
\newcite{das_etal_2014} improved on the prior state of the art in FrameNet argument span detection by using a joint model for all spans, and \cite{tackstrom_etal_2015} introduced a dynamic program for constrained inference of argument span detection and role labeling, and showed improvements in FrameNet and PropBank parsing. In both papers, frame predicate labeling was done separately, so that the joint model is more appropriately described as a model for Semantic Role Labeling (SRL).

In multi-task learning, in contrast, one or more models are learned that do different tasks separately, but the tasks share parameters or part of the architecture, thus enabling the models to learn better latent representations.
It has been shown that multi-task learning in deep models can improve results even for seemingly unrelated tasks \cite{collobert_weston_2008,sogaard_goldberg_2016}. Most recently, \newcite{2019t5} showed that large transformer-based models pre-trained on many tasks learn representations that are general enough to produce state of the art results in new tasks with relatively little fine-tuning.

In semantic parsing, multi-task learning has mostly focused on learning from multiple data sets with different semantic formalisms. \newcite{fitzgerald_etal_2015} trained a model for SRL on both FrameNet and PropBank and showed that the model achieves better results on FrameNet when the PropBank data is included. Similar results were reported by \newcite{Kshirsagar_etal_2015}. \newcite{peng_etal_2017} describe a model that jointly predicts multiple formalisms for each data point, and \newcite{peng_etal_2018} improve that model, which relies on a parallel corpus, by treating unannotated formalisms as latent structured variables.

Other work, motivated by the intuition that semantic parsing critically relies on syntax, and by the imperfection of syntactic parsers, has focused on multi-task learning of syntax and semantics. \newcite{swayamdipta_etal_2017} introduce a model architecture designed to produce both syntactic and semantic (FrameNet) parses and show that the semantic parse component performs better when it relies on the latent syntactic representation learned by the other component, without using the actual syntactic parse. \newcite{cai_lapata_2019} present a similar model for PropBank which utilizes two syntactic auxiliary tasks and achieves better results when incorporating this latent representation.


Our model follows much of this work in utilizing parameter sharing across tasks in order to leverage related tasks. It differs in that the related tasks, in our case, are all the sub-tasks of frame semantic parsing. Instead of leveraging data from related semantic formalisms, or related tasks such as syntactic parsing, we explore multi-task learning as an alternative to joint learning: sharing parameters across the various sub-tasks of frame semantic parsing to produce a better ultimate parse. Unlike previous work which focuses mainly on SRL, we learn to do frame predicate labeling with the same model so that no parsing sub-task is done externally.

\section{Background: Ontologies}
\label{sec:hector}
\label{section:Hector}

Before we describe our models and experiments, we briefly describe the three ontologies 
that will be used in our experiments. We use the term ``ontology" to mean an inventory of lexical meanings (which we will call ``frames'') of the language (English), related by semantic relations, in particular a specialization hierarchy.  Furthermore, a frame can have zero or more semantic roles (or arguments).

\subsection{PropBank}

PropBank is concerned with the semantics of verbs; each verb in a text is annotated with a single sense that is lemma-specific.  This sense is called a ``roleset" and corresponds to what we call a frame.  The inventory of roles is identical for all frames and divided into arguments and adjuncts, though the same argument role need not mean the same thing for different frames.

\subsection{FrameNet}

FrameNet \cite{framenet:book} has an orientation towards semantically defined situations, as opposed to the different senses of words.   These situations are called ``frames", which is also our terminology in this paper.  ``Frames represent story fragments, which serve to connect a group of words to a bundle of meanings" \cite{framenet:book}.  FrameNet has good (though not perfect) coverage of situations, and of the lexemes that describe them (primarily verbs for situations, but also some nouns).  Frames have frame elements, which correspond to what we call ``roles''.  The role labels are specific to a frame, and thus the inventory of role labels is much larger compared to PropBank.

\subsection{Hector}

We briefly introduce \emph{Hector}, a broad-domain frame ontology we created in-house for various NLP problems we are working on, such as question answering and knowledge extraction from unstructured text. 

Hector is a combination of FrameNet \cite{framenet:book}
and the New Oxford American Dictionary (NOAD) \cite{noad}, with the former providing broad coverage for multi-arity relations such as events/situations, while the latter being a comprehensive lexicon covering all parts of speech.  

We use FrameNet as our starting point because of its orientation towards situations rather than word meanings.
However, FrameNet's coverage for ``things" is insufficient.  For example, FrameNet has two meanings for the noun {\em ball}, namely \ecframe{Social\_event} and  \ecframe{Shapes} (as in {\em ball of thread}), but not the common sense of a sphere used in a game.  While FrameNet does not preclude being extended to include such meanings, we did not want to embark on such an extension on our own, and decided to combine FrameNet with an existing inventory of ``thing" meanings.

The WordNet \cite{fellbaum:1997} sense distinctions are too fine for our purposes, and the OntoNotes  \cite{hovy-etal-2006-ontonotes} sense groupings (which proposes coarser meanings) are insufficiently developed for nouns, for which we have sparse coverage from FrameNet. NOAD \cite{noad}, the New American Oxford Dictionary, on the other hand, contains all words of North American English, with sense distinctions at two levels, a coarse level and a finer level.  This coarser level corresponds roughly to the Ontonotes sense groupings, but has the advantage that it has been fully implemented for all nouns.  In addition, for nouns (but unfortunately only for nouns) there is a complete hierarchical upper ontology, which all nominal meanings are linked into.  The upper ontology has about 2,000 nodes.

Specifically, Hector consists of parts of FrameNet (for events and states), parts of NOAD (for things, and all adjectives and adverbs), and a small number of extensions we have made, almost exclusively to the FrameNet part.  In total, the ontology has around 95,000 frames, 725 of which have been derived from FrameNet, and includes 834 roles. 

\section{Transformers for Semantic Parsing}
\label{sec:models}
Recently, transformer-based deep learning models have achieved SOTA performance on various NLP benchmarks. Typically, a model for a downstream task (e.g. sentiment detection, textual entailment, question answering etc) is built by fine-tuning pre-trained language models (LMs), such as GPT2 \cite{gpt2} or BERT \cite{devlin2018bert}, on task-specific training data. Since the LMs have been pre-trained on a web-scale corpus, they implicitly capture a large amount of lexical, syntactic and semantic information about textual patterns. 

It seems natural to use pre-trained LMs for semantic parsing. We experiment with a few model formulations/architectures to tackle this problem and describe them in this section.

\subsection{Semantic Parsing as a Seq-to-Seq Generation Problem}
First, we consider modeling semantic parsing as a sequence-to-sequence generation problem, where the input is natural language text and the output is the corresponding frame representation encoded formally in text. 

For our initial experiments, we choose the target ontology to be Hector, since it is the largest ontology, and has both a rich concept hierarchy from NOAD and predicates with arguments/roles from FrameNet. One feature of Hector is that it has frame concepts corresponding to all contentful lexical tokens. Hence, while some frame semantic parsing problems such as FrameNet have a non-trivial step to detect a frame ``trigger" span in the text, we consider all nouns, verbs, adjectives, adverbs as frame triggers for Hector. Given the high accuracy for the part-of-speech (POS) tagging problem, we use an off-the-shelf POS tagger (Stanford) to detect triggers for the semantic parser.

\begin{table}[h]
    \centering
    \small
    \begin{tabular}{|p{3.5cm}|p{3.5cm}|}
        \toprule
        \textbf{Input} & \textbf{Output} \\\midrule
Two of the cast fainted and most of the rest * repaired * to the nearest bar. & repaired = \emph{Self motion} $\vert$ most of the rest = \emph{Self\_mover} $\vert$ to the nearest bar = \emph{Goal} $\vert$ \\ \hline
He blinked , taken aback by the * vigour * of her outburst. & vigour = \emph{Dynamism} $\vert$ of her outburst = \emph{Action} $\vert$ \\ \hline
The rain * dripped * down his neck. & dripped = \emph{Fluidic motion} $\vert$ The rain = \emph{Fluid} $\vert$ down his neck = \emph{Path} $\vert$ \\ \hline
She * adored * shopping for bargains and street markets and would have got on well with Cherry. & adored = \emph{Experiencer focus} $\vert$ She = \emph{Experiencer} $\vert$ shopping for bargains and street markets = \emph{Content} $\vert$ \\ \hline
He * cleared * his throat as the young man looked up. & cleared =  \emph{Emptying} $\vert$ He = \emph{Agent} $\vert$ his throat = \emph{Source} $\vert$ \\
        \bottomrule
    \end{tabular}
    \caption{Training data format for the generative seq2seq approach}
    \label{table:trainingBaseline}
\end{table}

Given a sentence and a trigger-marked span, we use a generative model to produce the corresponding frame interpretation for the trigger. As shown in Table \ref{table:trainingBaseline}, we mark the trigger span in the text with asterisks, and use a \emph{key = value} format in the frame representation output, where each key is a text span and its value is the corresponding frame or role label. 

Advantages of this representation include the ability to specify overlapping argument spans, which cannot be done using the standard BIO formulation typically used for sequence labeling, and the fact that all the elements of the full frame interpretation are generated jointly. 

By using FrameNet text annotations and example sentences in NOAD, we were able to collect 1.4M frame annotated sentences in total. We then split the data into training, validation and test sets using a 80/10/10 split done randomly. 

\begin{table}[h]
    \centering
    \small
    \begin{tabular}{|l|p{0.9cm}|p{0.8cm}|p{0.7cm}|p{0.5cm}|}
        \toprule
        \textbf{Generative Model} & \textbf{Frame Accuracy} & \textbf{Role Precision}  & \textbf{Role Recall} & \textbf{Role F1} \\\midrule
        GPT2-small (117M) &  77\% & 60\% & 59\% & 59\% \\ 
        GPT2-medium (345M) &  79\% & 73\% & 71\% & 72\% \\ 
        GPT2-large (770M) &  82\% & 77\% & 76\% & 77\% \\ 
        T5-small (120M) &  82\% & 77\% & 81\% & 79\% \\ 
        \textbf{T5-base (440M)} & \textbf{87\%} & \textbf{81\%} & \textbf{83\%} & \textbf{82\%} \\
        \bottomrule
    \end{tabular}
    \caption{Results for semantic parsing as a fully generative translation problem on the test set}
    \label{table:baselineResults}
\end{table}

We tested this approach using both a language model (GPT2) and an encoder-decoder model (T5). We also experimented with various model sizes (GPT2 small-large, T5 small-base)
to determine how model size/capacity affects performance. All models were trained with the same hyper-parameters (5 epochs, 1e-3 learning rate, 64 batch size).

The results on the test set are shown in Table \ref{table:baselineResults}. The metrics are the classification accuracy to determine the frame sense, and the micro precision, recall and F1 scores for role detection (this corresponds to the {\bf {\microsoftmatch}} metric described in  Section \ref{sec:evalmetrics} on our Evaluation Metrics). 

Not surprisingly, a larger model size results in improved performance for frame classification and role detection (at the cost of additional memory and training time for the extra parameters). However, an interesting result is the performance of T5-small, which beats that of GPT2-large, even though the former has fewer than one fifth of the latter's parameters. This clearly indicates the importance of selecting an appropriate model architecture for the task. 

GPT2, being a language model, does not have distinct notions of input and output sequences, and instead learns to predict the next token given a prior text sequence, employing a stacked encoder to represent the prior text. While this model can be used for our problem of semantic parsing by feeding it a full training example with a clear separator (e.g. tab) between the input and output, during training, the model still attempts to predict elements of the input, which is unnecessary for our problem. On the other hand, T5 is based on an encoder-decoder architecture, and does have a clear notion of input/output sequences, using a stacked encoder to represent the input and a stacked decoder to represent the output.  During training, the model learns to only decode the output sequence conditioned on the full input. This makes it a much better fit for our problem, as the results demonstrate.  

\subsection{Semantic Parsing as a Multi-task Mixed Decoding Problem}

As mentioned earlier, semantic parsing is often decomposed into a series of tasks - trigger detection in the text, frame (predicate) classification given a trigger, and role detection given a trigger and frame label. In our work, we assume that the trigger is provided (which is trivial in the Hector frame ontology case, since all contentful tokens are triggers).

\begin{table}[h]
    \centering
    \small
    \begin{tabular}{|p{5cm}|p{2cm}|}
        \toprule
        \textbf{Input} & \textbf{Output} \\\midrule
FRAME: 0 Two  1 of  2 the 3 cast 4 fainted 5 and 6 most 7 of 8 the 9 rest 10 * repaired * 11 to 12 the 13 nearest 14 bar 15 . & Self motion \\ 
ARGS for Self motion: 0 Two  1 of  2 the 3 cast 4 fainted 5 and 6 most 7 of 8 the 9 rest 10 * repaired * 11 to 12 the 13 nearest 14 bar 15 . & Self\_mover = 6-9  $\vert$ Goal = 11-14 $\vert$  \\ \hline

FRAME: 0 He 1 blinked 2 , 3 taken 4 aback 5 by 6 the 7 * vigour * 8 of 9 her 10 outburst. & Dynamism \\
ARGS for Dynamism: 0 He 1 blinked 2 , 3 taken 4 aback 5 by 6 the 7 * vigour * 8 of 9 her 10 outburst. & Action = 8-10 $\vert$ \\ \hline

FRAME: 0 The rain 1 * dripped * 2 down 3 his 4 neck. & Fluidic motion \\
ARGS for Fluidic motion: 0 The rain 1 * dripped * 2 down 3 his 4 neck. & Fluid = 0-1 $\vert$ Path = 2-4 $\vert$ \\
        \bottomrule
    \end{tabular}
    \caption{Training data format for 2-task frame parsing: Frame classification, and Argument detection conditioned on the frame.}
    \label{table:trainingMultiTask}
\end{table}

For this multi-task formulation, we encode the problem as shown in Table \ref{table:trainingMultiTask}, where a single training example is split into two examples, one per task. The task itself is specified as a prefix ``command" to the input context ("FRAME:", "ARGS for \emph{frame-label}:" are the task commands). Note that the input context (i.e. sentence with the trigger span marked within asterisks) is the same for both tasks. 

One additional modification we made, compared to the earlier seq2seq format, is to add token indices before each token in the input, and specify the argument spans in the output in terms of token index ranges. The intuition was to leverage the positionally-aware representations captured by transformer models, and to also shorten the generative text output, which improves decoding speed. Also, by using exact token indices, we avoid any ambiguity in the final frame representation, which would arise if an argument text span appears multiple times in the sentence.

\begin{figure}[h]
\centering
\includegraphics[width=1\columnwidth]{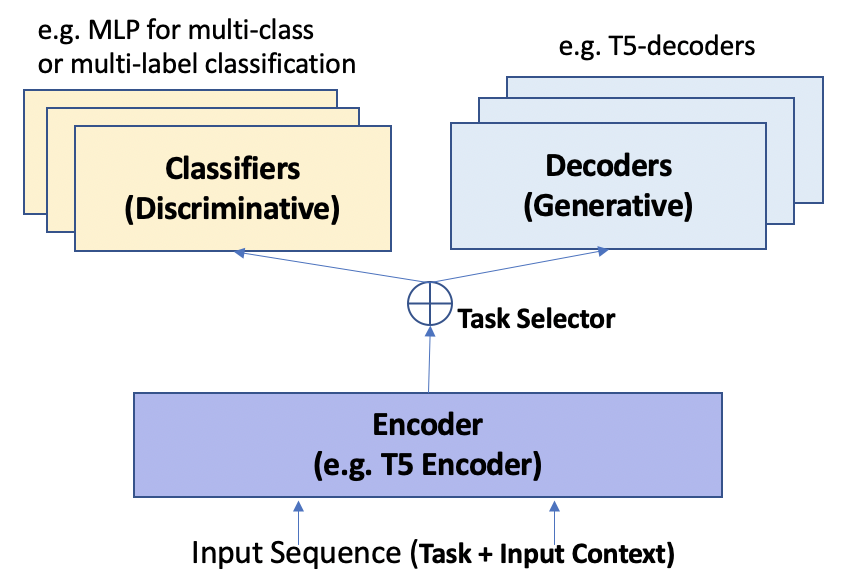}
\caption{Multi-task model for semantic parsing}
\label{fig:spindle-arch}
\end{figure}

To appropriately utilize this training data format, we design a general purpose multi-task encoder-decoder architecture (see Figure \ref{fig:spindle-arch}), where the encoder layer is shared among the various tasks, with different classifiers and decoders used depending on the task type. 

The training routine has a specialized batch collator and sampler which partitions the training data so that each batch sent to the model during training is for a single task. Batches are randomly sampled in a round-robin fashion among the tasks. The loss function is a weighted sum of the individual task loss functions. We use a simple loss-balancing algorithm to dynamically update the task weights to ensure that the losses (in particular, the exponential moving average of the loss) for the various tasks are on the same scale. As part of future work, we plan on adding more sophisticated routines for adjusting weights based on individual task learning rates (e.g. see Gradient Norm \cite{GradientNorm}).

At prediction time, we do frame-sense classification before doing argument detection since argument detection is conditioned on the predicted frame, and specified as part of the task command. Finally, we combine the results of the two steps to produce the final frame interpretation (as shown in the examples in Table \ref{table:spindleExamples}). 

We ran an experiment using this multi-task model and the format in Table \ref{table:trainingMultiTask} on the same train/dev/test split, and with the same hyperparameters. We used the T5-base pre-trained encoder as the shared encoder for both steps. At decoding time, we used a single layer MLP (input-size: number of dimensions in encoder output, output-size: number of classes), with the standard categorical cross-entropy loss to predict frame predicates, and the generative T5-base decoder to produce argument spans and labels. The results are shown in Table \ref{table:multitaskResults}, and we see a 3\% improvement on classification accuracy, and a 2\% improvement on Argument F1, compared to the fully generative single task approach.

\begin{table}[h]
    \centering
    \small
    \begin{tabular}{|l|p{0.9cm}|p{0.8cm}|p{0.7cm}|p{0.5cm}|}
        \toprule
        \textbf{Model} & \textbf{Frame Accuracy} & \textbf{Role Precision}  & \textbf{Role Recall} & \textbf{Role F1} \\\midrule
        Generative & 87\% & 81\% & 83\% & 82\% \\
         \textbf{Multi-task} & \textbf{90\%} & \textbf{85\%} & \textbf{83\%} & \textbf{84\%} \\
        \bottomrule
    \end{tabular}
    \caption{Multi-task model results for semantic parsing. 
    }
    \label{table:multitaskResults}
\end{table}

\subsection{Discussion}

As our experimental results show, the multi-task model significantly outperforms simpler models on the Hector test set, for both frame classification and role detection. Looking at the examples in Table \ref{table:spindleExamples}, we can see that the model is able to correctly identify the frame sense for ambiguous verbs (such as {\em get}) based on the surrounding context. It can also detect general role arguments such as \emph{explanation} even though this role is not specifically defined for the corresponding frame (\ecframe{Ingestion}), and is instead learned from other frame annotated examples. 
A particularly interesting example is the last row of Table \ref{table:spindleExamples} where the model uses the syntax around the verb "sneeze" to infer that the implied frame in this context is \ecframe{removing} the napkin off the table. 

One of the nice features of the multi-task model design is that by treating frame prediction as a classification problem, we are able to get reliable, normalized (0-1) confidences from the classifier, which can be used as probability estimates for the prediction. Contrast this with a purely generative model where the confidences (even if normalized to be in the 0-1 range) are not comparable across texts and vary depending on the input text content/length. 

On the flip side, an advantage of the purely generative approach (compared to using classifiers) is that it is straightforward to add new frames/roles to the vocabulary without changing the model architecture. This is because the model does not have a pre-defined list of frames/roles (which the classifier needs), and instead directly generates vocabulary tokens. Moreover, tokenizers for models such as T5 typically split words to the sub-token level, and in our experience rarely produce OOV terms for our problem. There are ways to alleviate this issue with the classifier approach, e.g. by using `spare' classes which remain unused during training, and are assigned to new frames/roles as needed. 

\subsection{Next Steps}

We plan to use a multi-label classifier for frame prediction, to exploit the frame hierarchy in Hector. This would also let the model predict multiple non-disjoint frame senses when applicable, such as the {\em sneeze} example in Table \ref{table:spindleExamples}, where it is arguable that both \ecframe{removing} and \ecframe{makeNoise} (the more direct meaning of sneeze) are applicable in the given context. We also plan to split argument detection into separate tasks of span detection and role label classification (as traditionally done) in order to derive role-confidences which can be factored into the final frame interpretation confidence. However, it is an empirical question whether performance would drop due to the cascading error problem in pipeline approaches.

Finally, we plan to exploit the multi-task encoder-decoder architecture to do a form of \textbf{transfer learning} by adding related (conceptually similar) tasks to the model and seeing if performance improves across all tasks. For example, we intend to train a model that does semantic parsing over all the ontologies (Hector, FrameNet and PropBank) together by representing them as separate tasks, yet still sharing a common encoder.

\section{Experiments on Standard Data Sets}
\label{sec:benchmark_experiments}
We ran our transformer-based models on the FrameNet 1.7 and the PropBank SRL CoNLL 2012 benchmarks, and compared performance with recent state of the art systems. Before we dive into the results, we describe our evaluation metrics.

\subsection{Evaluation Metrics}

\label{sec:evalmetrics}
For frame prediction, we use the standard classification accuracy as our metric. For argument (role) detection, we report results using 3 different metrics:
\begin{itemize}[noitemsep]
    \item  {\bf {\exactmatch}}: We require both the argument span and the role label to match exactly.  We perform a recall-precision analysis over all (argument-label,span) pair instances in the test set.
    \item {\bf {\softmatch}}: We split argument spans into individual tokens and consider pairs of (argument-label, token) in the system output and gold data.  We perform a recall-precision analysis on these pairs for a single role instance, and then average the recall and precision over all role instances in the test set and calculate the F1 measure from these averages.
    \item {\bf {\microsoftmatch}}: Similar to {\softmatch}, except we 
    compute recall and precision across all instances of pairs of (argument-label,token) in the test set.  (This amounts to micro-averaging.)
\end{itemize}

FrameNet has its own evaluation script developed for the 2007 SemEval Shared Task 19  \cite{baker-etal-2007-semeval}.  The version that has been used starting with \newcite{das_etal_2014} uses the ``labels-only" option, which requires exact match on role spans (and role labels) and thus corresponds to our strictest option {\bf{\exactmatch}}.  In contrast, the matching with the predicted frame is more lenient, as the FrameNet evaluation script allows for partial credit if the predicted frame is related to the gold frame through frame-to-frame relations.  While we acknowledge the interest in giving partial credit for closely related frames, we find that frame-to-frame relations do not represent a uniform notion of semantic distance, and we prefer to keep a strict identity.  As a result, our {\exactmatch} results  are most comparable to previous published results, but they are stricter.  To provide a calibration, we evaluate the best published system \cite{swayamdipta_etal_2017} using our three metrics.

Our {\softmatch} and {\microsoftmatch} give partial credit for predicted role spans if they overlap with the gold span (but only if the predicted label is correct).  Recall and precision are calculated on a token basis, while for  {\exactmatch} they are calculated on a span basis.  As a result, while we expect  {\softmatch} and {\microsoftmatch} results to be better than  {\exactmatch} results, this is not necessarily the case.  Furthermore,  {\softmatch} gives equal weight to each role instance, so that, in comparison with {\microsoftmatch}, errors in long spans are not weighted as highly.  In contrast, a system that has trouble finding span boundaries especially in long spans will perform worse on {\microsoftmatch}  than on  {\softmatch}.

\subsection{Framenet 1.7 Evaluation}

We used the FN 1.7 evaluation dataset, 
using gold triggers.
We use the prior SOTA system {\sesame} \cite{swayamdipta_etal_2017} as our point of comparison. Results for frame classification are in Table~\ref{table:fn1.7-class} and role detection in Table~\ref{tab:fn1.7-args}.

Considering the frame detection task, we see that our two systems, the straightforward seq2seq approach (titled ``{\fullgen}" in the table) and the multi-task approach (``{\multitask}"), perform about equally on the test set, and only minimally better than the state of the art system, {\sesame}.  (We have not performed significance testing.)  Also, the difference in performance between {\sesame} and our systems is more pronounced in the dev set, suggesting that we tuned on the dev set more than did {\sesame}.

In contrast, for role prediction (Table~\ref{tab:fn1.7-args}), we see great differences.  Concentrating first on the {\exactmatch} metric, our two models both outperform the state of the art model, with substantial increases of at least 12\% absolute F1.  Furthermore, {\multitask} again improves on {\fullgen} with improvements of at 1-5\% absolute F1.  Interestingly, {\sesame} generally has a higher precision than recall, while our systems perform roughly equally on the two measures.   Comparing the dev and test sets, we see again that {\sesame} improves on test compared to dev, while our systems show decreased performance.  Again, we assume this is because we used the dev set more extensively in tuning our system, and we therefore slightly overfitted to dev.  We now compare results with gold predicates to results on predicted frames (i.e., to results on an end-to-end semantic parser, which is the most realistic way of measuring performance).  As expected, all systems perform better with gold frames, but {\sesame} has a slightly lower decrease than our systems.


\begin{table}[bh]
    \centering
    \small
    \begin{tabular}{|l|c|c|}
        \toprule
        \textbf{Model} & \textbf{Frame Acc. Dev} & \textbf{Frame Acc. Test} \\\midrule
       {\sesame} & 88.9\% & 86.5\% \\ 
       {\fullgen} & 90.5\% & 87\% \\
    {\multitask} & \textbf{90.5\%} & \textbf{87.5\%} \\
        \bottomrule
    \end{tabular}
    \caption{Frame Detection results for FN 1.7}
    \label{table:fn1.7-class}
\end{table}

\begin{table*}[bht]
    \centering
    \small
    \begin{tabular}{|p{5em}p{0.1\linewidth}|ccc|ccc|ccc|ccc|}
        \hline
         \bf Metric & \bf Model & \multicolumn{3}{c|}{\bf GOLD-dev}  &  \multicolumn{3}{c|}{\bf GOLD-test} & \multicolumn{3}{c|}{\bf PRED-dev} &  \multicolumn{3}{c|}{\bf PRED-test} \\ 
         & & P & R & F1 & P & R & F1 & P & R & F1 & P & R & F1 \\ \hline

       \multirow{3}{*}{\exactmatch} & {\bf Sesame} & 0.60 & 0.51 & 0.55 & 0.62 & 0.55 & 0.58 & 0.55 & 0.47 & 0.51 & 0.57 & 0.49 & 0.52 \\ 
        & {\fullgen} & 0.71 & 0.73 & 0.72 & 0.71 &  0.73 & 0.72 & 0.65 & 0.66 & 0.66 & 0.63 & 0.65 & 0.64 \\
        & {\multitask} & 0.77 & 0.77 & \bf 0.77 & 0.75 & 0.76 & \bf 0.76 & 0.71 & 0.70 & \bf 0.71 & 0.66 & 0.67 & \bf 0.66 \\ \hline
        
        \multirow{3}{*}{\softmatch} & \bf Sesame & 0.71 & 0.61 & 0.66 & 0.71 & 0.64 & 0.67 & 0.64 & 0.56 & 0.60 & 0.63 & 0.56 & 0.59 \\ 
        & {\fullgen} & 0.80 & 0.81 & 0.80 & 0.78 & 0.80 & 0.79 & 0.73 & 0.74 & 0.73 & 0.69 & 0.71 & 0.70 \\
        & {\multitask} & 0.83 & 0.82 & \bf 0.82 & 0.80 & 0.82 & \bf 0.81 & 0.75 & 0.75 & \bf 0.75 & 0.71 & 0.72 & \bf 0.71 \\
   \hline
   
          \multirow{3}{*}{\parbox{5em}{\microsoftmatch}} & \bf Sesame & 0.66 & 0.51 & 0.58 & 0.62 & 0.52 & 0.57 & 0.59 & 0.45 & 0.51 & 0.53 & 0.45 & 0.59 \\ 
       & {\fullgen} & 0.8 & 0.77 & 0.79 & 0.74 & 0.76 & 0.75 & 0.72 & 0.70 & 0.71 & 0.64 & 0.66 & 0.65 \\
       & {\multitask} & 0.82 & 0.80 & \bf 0.81 & 0.75 & 0.80 & \bf 0.77 & 0.75 & 0.73 & \bf 0.74 & 0.65 & 0.69 & \bf 0.67 \\ 
       
         \hline   
    \end{tabular}
    \caption{Argument Detection Results for FN 1.7; ``GOLD" refers to gold wordsenses, while ``PRED" refers to predicted word senses; Sesame is the state-of-the-art system of \citet{swayamdipta_etal_2017} ({\sesame}) evaluated using our metrics, {\fullgen} is our sequence-to-sequence generative system, and {\multitask} is our multi-task approach}
    \label{tab:fn1.7-args}
\end{table*}

We now consider the results using all three of our metrics.  As expected, all systems on all measures perform better using {\softmatch} compared to {\exactmatch}.  However, when we compare  {\microsoftmatch} with {\exactmatch}, we see that {\sesame} does not show improved results, while our systems have equal or better results.  We take this to mean that {\sesame}, in contrast to our systems, has specific problems with longer role spans.  Errors in long role spans are more heavily penalized in {\microsoftmatch}, as false tokens in the predicted spans are counted globally across all predicted roles.  In contrast, for {\softmatch}, each instance of a predicted role span has equal weight.  Given the close relation between syntactic structure and semantic role spans, we conclude that our systems, without modeling syntax explicitly at all, are better at implicitly modeling syntax than {\sesame}, which uses a ``syntactic scaffold" during training (but not decoding).

\subsection{PropBank CoNLL 2012 Evaluation}
We compare our models to two recently published systems that have achieved SOTA results for PropBank SRL parsing: \cite{he-etal-2018-jointly} and \cite{LiSRL2019}. The benchmark we use is CoNLL 2012 \cite{pradhan-etal-2013-towards} which is based on Ontonotes 5.0 and also includes nominal predicates. Since we did not have access to the third party systems, we use the official CoNLL scripts for evaluation, as a means of comparison. Note that evaluation is micro-averaged F1 for correctly predicting (predicate, argument span, label) tuples, which corresponds to our {\exactmatch} metric. 

\begin{table}[H]
    \centering
    \small
    \begin{tabular}{|l|p{0.9cm}|p{0.8cm}|}
        \toprule
        \textbf{Model} & \textbf{Dev} & \textbf{Test} \\\midrule
        \cite{he-etal-2018-jointly} & 83\% & 82.9\% \\
        \cite{LiSRL2019} & - & 83.1\% \\
        Full-Gen & 82.4\% & 82.3\% \\
         \textbf{Multi-task} & \textbf{83.4\%} & \textbf{83.7\%} \\
        \bottomrule
    \end{tabular}
    \caption{Results for end-to-end Propbank SRL parsing on CoNLL 2012. 
    }
    \label{table:propbankResults}
\end{table}

The evaluation we perform is on end-to-end SRL prediction (i.e. predicting the verb frame and the arguments), which is the most realistic use-case. The results are shown in Table \ref{table:propbankResults}. The multi-task model performed the best at this task, while the fully generative approach did not beat the prior state of the art (though was not far off). 

\section{Error Analysis}

We performed an error analysis on the FrameNet predictions of the {\fullgen} model.  We randomly selected 60 instances of predicted semantic analyses (trigger and roles) from the dev set.  We analyzed the data that is used for the PRED-dev columns in Table~\ref{tab:fn1.7-args}.   Some of our errors have wrong predicted frames.  In this case, all predicted roles count as false positives (FPs), and all gold roles as false negatives (FNs).  If the frame is predicted correctly, we look at the roles in more detail; note that at least one role is incorrect, but many roles may be correct.  In total, there were 167 false FPs and FNs across the 60 predictions.  By tabulating them across the instances, our error analysis follows the spirit of our {\microsoftmatch} metric.  The results are summarized in Table~\ref{tab:erroranalysis}.

\begin{table}[H]
    \centering
    \small
    \begin{tabular}{|l||c|c|c|c|}
        \hline
        \textbf{Error Type} & \textbf{FN} & \textbf{FP} & \textbf{Total} & \bf \%\\
        \hline
        \textbf{Gold error}  & 7 & 12 & 19 & 11.4 \\
        \textbf{Gold questionable}  & 16 & 14 & 30 & 18.0 \\  
        \textbf{Frame error}  & 34 & 37 & 71 & 42.5 \\  
        \textbf{Role error} & 25 & 22 &  47 & 28.1 \\
        \hline
        \textbf{Total} & 82 & 85 &  167 & 100 \\
        \hline        
    \end{tabular}
    \caption{Error Analysis for FrameNet parsing using Full-Gen}
    \label{tab:erroranalysis}
\end{table}

In any task as complex as semantic annotation, we expect to find some annotation errors.  We disagreed with two frame (predicate) choices.  Furthermore, we had four cases of a wrong choice of role for a correct span, and one case of a correct role but with a wrong span.  Finally, two gold analyses missed a role.  We recognize that despite our careful reading of the frame definitions in FrameNet, we may have misunderstood the intention of a frame or role.  

We also found cases in which gold and predicted differed, but we could not determine that one of the two was clearly incorrect.   These fall into two cases for frames.  In the first category, the two frames seem to be both true at once (\ecframe{Sounds}/\ecframe{Sensation}).  In the second category, it is hard to tell what is meant in context: for example, in {\em U.S. intelligence analysts}, does {\em intelligence} evoke an \ecframe{Organization} (gold) or does it refer to \ecframe{Information} (predicted)?  We found five frames that are questionable (15 FNs and FPs), and 15 instances of roles where either the span or the FE choice is questionable for various reasons.

18 frames (of 60) were predicted wrong in this set of errorful analyses.  \ocrcomment{Need to comment on the fact that this seems larger than expected from comparing GOLD-dev to PREF-dev (.79 to .73).  Wait for updated results from Adi before commenting.}. These entailed 71 FPs and FNs.  There is unfortunately no particularly clear pattern that arises for the role errors for correctly predicted frames.   23\% of the 47 role errors are roles missed altogether (FNs), 17\% are wrong additional predicted roles (FPs), 30\% are right span but wrong role name (FNs and FPs), and 30\% are right role name but wrong span (FNs and FPs).  

Overall, we see that {\fullgen} does not systematically over- or under-generate roles.  The major source of errors is the frame detection step (choosing the frame), suggesting the focus of further improvement.  The nearly 30\% of cases in which the gold standard is wrong or questionable is as expected for semantic parsing.
 
\section{Conclusion}
\label{sec:conclusion}
We explored the use of pre-trained transformer-based models for frame semantic parsing. In preliminary experiments, we discovered that encoder-decoder architectures significantly outperform language model architectures, even when they contain far fewer parameters. We show that a seq2seq generative formulation using the T5-base model  handily beats the previous state-of-the-art in the FrameNet 1.7 benchmark, which highlights the ability of pre-trained transformers to implicitly capture a large amount of lexical, syntactic and semantic information. Finally, we introduce a multi-task variant which employs specialized decoders (and parameter sharing) for the various subtasks of the semantic parsing task, and achieves the highest performance on both the FrameNet and PropBank SRL datasets.

\section*{Acknowledgments}

We thank Swabha Swayamdipta for supplying her system's output, and for extensive help with the FrameNet evaluation script.

\newpage

\bibliography{anthology,nonacl, semparser}
\bibliographystyle{acl_natbib}

\end{document}